\documentclass[proc]{edpsmath}

\usepackage{amssymb,amsmath,mathrsfs,amsfonts,graphicx,epsf,color}

\usepackage{todonotes}
\usepackage{subcaption}

\newcommand{\R}{\mathbb{R}}
\newcommand{\PTR}{PT\R^2}

\newcommand{\bqn}{\begin{eqnarray}}
\newcommand{\eqn}{\end{eqnarray}}

\newcommand{\ba}[1]{\begin{array}{#1}}
\newcommand{\ea}{\end{array}}
\renewcommand{\th}{\theta}

\newcommand{\eqnn}{\nonumber\end{eqnarray}}
\newcommand{\eqnl}[1]{\label{#1}\end{eqnarray}}

\newcommand{\bdeff}{\begin{definition}}
\newcommand{\edeff}{\end{definition}}

\newcommand{\bd}{\begin{description}}
\newcommand{\ed}{\end{description}}

\newtheorem{theorem}{Theorem}

\newtheorem{proposition}[theorem]{Proposition}
\newtheorem{definition}[theorem]{Definition}
\newtheorem{remark}[theorem]{Remark}
\newcommand{\brem}{\begin{remark}}
\newcommand{\erem}{\end{remark}}

\renewcommand{\r}[1]{(\ref{#1})}

\title{Cortical-inspired image reconstruction via sub-Riemannian geometry and hypoelliptic diffusion}
\runningtitle{Cortical-inspired image reconstruction}

\begin{document}

\author{U.\ Boscain}
  \address{CNRS, LJLL, Universit\'e Pierre et Marie Curie, Paris, France}
  \secondaddress{\'Equipe INRIA Paris CAGE}
\author{R.\ Chertovskih}
  \address{SYSTEC, FEUP, University of Porto, Portugal}
  \secondaddress{Samara University, Russia}
\author{J.P Gauthier}
  \address{LSIS, Universit\'e de Toulon, France}
\author{D.\ Prandi}
  \address{CRNS, L2S, CentraleSup\'elec, Gif-sur-Yvette, France}
\author{A.\ Remizov}
  \address{CMAP, \'Ecole Polytechnique, Palaiseau, France}

\begin{abstract}
   In this paper we review several algorithms for image inpainting based on the hypoelliptic diffusion naturally associated with a mathematical model of the primary visual cortex. In particular, we present one algorithm that does not exploit the information of where the image is corrupted, and others that do it. While the first algorithm is able to reconstruct only images that our visual system is still capable of recognize, we show that those of the second type completely transcend such limitation providing reconstructions at the state-of-the-art in image inpainting. This can be interpreted as a validation of the fact that our visual cortex actually encodes the first type of algorithm.
\end{abstract}



\maketitle

\section{Introduction}

Since the founding works of Petitot, Citti and Sarti \cite{petitot,citti-sarti}, where the foundations of the mathematical model (known as Citti-Petitot-Sarti model, CPS model in short) of the human primary visual cortex V1 have been posed, many authors worked on develop and apply these ideas to image processing and computer vision \cite{Duits2010,Duits2010a,Sanguinetti,Bredies13,Hladky}. Indeed, the main concept at the basis of the CPS model, i.e., that images in our brains are processed via a redundant representation taking into account local features as local orientation, presents a somewhat new framework in { which many} tasks can be strongly simplified. 

\begin{figure}
  \centering\raisebox{-0.5\height}{
  \begin{minipage}[b]{.3\linewidth}
    \centering\raisebox{-0.5\height}{\includegraphics[width=.9\textwidth]{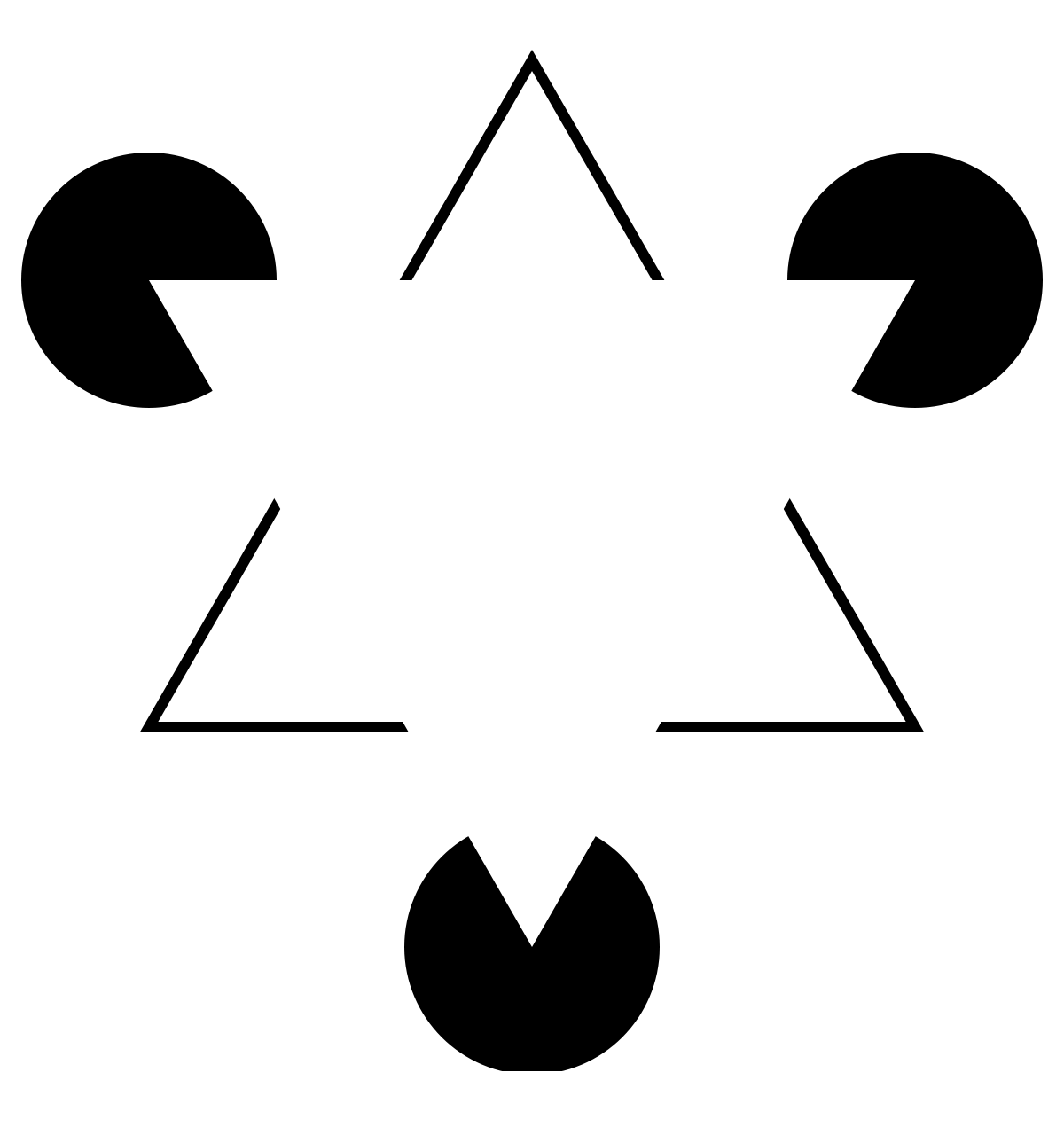}}
    \subcaption{Modal completion (Kanisza triangle).}
  \end{minipage}
  }
  \qquad
  \centering\raisebox{-0.5\height}{
  \begin{minipage}[b]{.3\linewidth}
    \includegraphics[width=.9\textwidth]{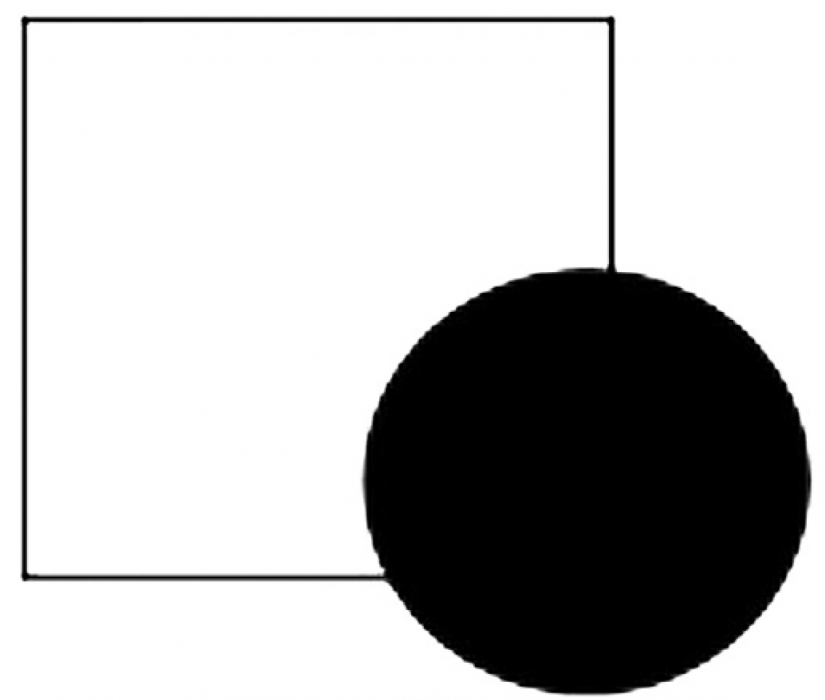}
    \vspace{1em}
    \subcaption{Amodal completion.}
  \end{minipage}
  }
  \caption{The two different neurophysiological types of contour completion.}
  \label{fig:modal}
\end{figure}

In view of the many well-studied examples of contour completion operated by the human brain (see Figure~\ref{fig:modal}), one of the main problems that was attacked from this point of view has been that of image reconstruction, also known as image inpainting \cite{Bertalmio2000,Masnou2002,Chan2002,Criminisi2004a}.
In this survey, we present the results we obtained in a series of works \cite{Boscain2012a, Remizov2013, Boscain2014, ahe, gros-papier} in this context. The paper is divided in three parts, corresponding to different methods and purposes of image reconstruction.

In the first part, Section~\ref{s-petitot}, we introduce the geometry of the problem by describing the Citti-Petitot-Sarti model for reconstruction of curves, with some new ingredients introduced in \cite{Boscain2012a}. Although this simplified model is not very effective to reconstruct images partially corrupted (reconstructing an image level curve by level curve is a hopeless problem), it permits to understand easily the sub-Riemannian structure of V1, which will allow to introduce the hypoelliptic diffusion.

Indeed, in Section~\ref{s-hypo}, we present an image inpainting algorithm obtained by building on the curve reconstruction methods of the first part. In particular, we show that, via stochastic considerations, the latter can be extended to image reconstruction by considering the hypoelliptic diffusion associated with the a sub-Riemannian structure of V1. Although not very efficient, this method allows to reconstruct images with small corruptions. The main interest of this technique is to simulate the reconstruction of images by the visual cortex and, as we will show in the next section, to produce very effective image inpainting algorithms when coupled with other techniques. Moreover, thanks to the exploitation of non-commutative harmonic analysis techniques, the final algorithm that we present has the twofold advantage of, on one side, being very fast and easy to parallelize, and on the other, of not needing nor exploiting any information on the location and shape of the corruption. 

Finally, in the third and last part, Section~\ref{s-corr}, we collect some results detailing how to ameliorate the above hypoelliptic diffusion method by taking into account the location of the corruption. In particular, we show how these methods allow to obtain image reconstructions of highly corrupted images (e.g.\ with up to 97\% of pixel missing), placing themselves at the same level as state-of-the-art inpainting methods. 

We conclude this brief introduction by observing that, experimentally, the threshold of the maximal amount of corruption that the pure hypoelliptic diffusion method can manage looks very close to the threshold of corruption above which our brain can recognize the underlying image. This fact could be seen as a validation of the CPS model with pure hypoelliptic diffusion.
  On the other hand, the methods presented in the last part completely transcend such problem, thus suggesting that, although based on neurophysiological ideas, they do not reflect any real mechanism of V1.


\section{The sub-Riemannian model for curve reconstructions}
\label{s-petitot}

In this section, we recall a model describing how the human visual cortex V1 reconstructs curves which are partially hidden or corrupted. 
The model we present here was initially due to Petitot \cite{petitot,petitot-libro}. It is based on previous work by Hubel-Wiesel \cite{hubel} and Hoffman \cite{hoffman}, then it was refined by Citti \emph{et al.} \cite{citti-sarti,Citti2016}, Duits \emph{et al.} \cite{Duits2008,Duits2010,Duits2010a,Duits2014}.
and by the authors of the present paper in \cite{Remizov2013,Boscain2012a,Boscain2014,ahe}. It was also studied by Hladky and Pauls in \cite{Hladky}.

In a simplified model\footnote{For example, in this model we do not take into account the fact that the continuous space of stimuli is implemented via a discrete set of neurons.} (see \cite[p. 79]{petitot-libro}), neurons of V1 are grouped into {\it orientation columns}, each of them being sensitive to visual stimuli at a given point  of the retina and for a given direction\footnote{Geometers call ``directions'' (angles modulo $\pi$) what neurophysiologists call ``orientations''.} on it. The retina is modeled by the real plane, i.e. each point is represented by $(x,y)\in\R^2$, while the directions at a given point are modeled by the projective line, i.e. $\theta\in P^1$. Hence, the primary visual cortex V1 is modeled by the so called {\it projective tangent bundle} $\PTR:=\R^2\times P^1$. From a neurological point of view, orientation columns are in turn grouped into {\it hypercolumns}, each of them being sensitive to stimuli at a given point $(x,y)$ with any direction. In the same hypercolumn, relative to a point $(x,y)$ of the plane, we also find neurons that are sensitive to other stimuli properties, like colors, displacement directions, etc...  In this paper, we  focus only on directions and therefore  each hypercolumn is represented by a fiber $P^1$ of the bundle $\PTR$.

\begin{figure}
  \includegraphics[width = .7\textwidth]{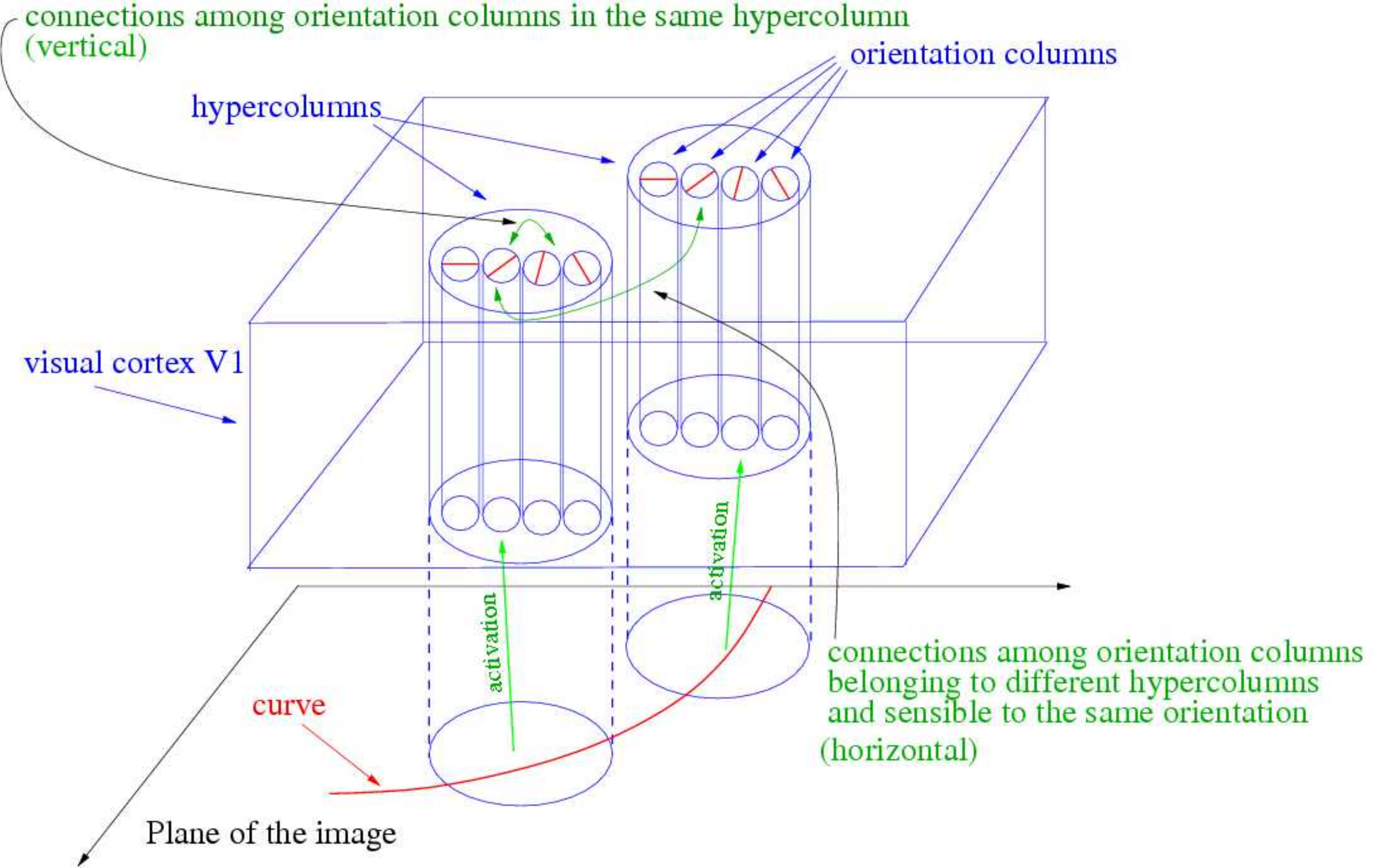}
  \caption{A scheme of the primary visual cortex V1.}
  \label{fig:f-hyper-bis}
\end{figure}

Orientation columns are connected between them in two different ways. The first kind of connections are the  {\it vertical} (inhibitory) ones, which connect orientation columns belonging to the same hypercolumn and sensible to similar directions. The second kind of connections are the  {\it horizontal} (excitatory) connections, which connect neurons belonging to different (but not too far) hypercolumns and sensible to the same directions. (See Figure~\ref{fig:f-hyper-bis}.) These two types of connections are represented by the following vector fields on $PT\mathbb R^2$:
\begin{equation}
  X(x,y,\theta) = \left(\begin{array}{c} \cos\theta \\ \sin\theta\\0 \end{array}\right) \quad\text{ and }\quad
  \Theta(x,y,\theta) = \left(\begin{array}{c} 0 \\ 0\\ \beta \end{array}\right).
\end{equation}
The parameter $\beta>0$ is introduced here to fix the relative weight of the horizontal and vertical connections, which have different physical dimensions. 

In other words, when V1 detects a (regular enough) planar curve $\gamma:[0,T]\to\R^2$, $\gamma(t)=(x(t),y(t))$, it computes a ``lifted curve''  $\Gamma:[0,T]\to\PTR$, $\Gamma(t) = (x(t),y(t),\theta(t))$, by including a new variable $\theta(\cdot):[0,T]\to P^1$ which satisfies:
\begin{equation}\label{eq-contrSR}
  \frac{d}{dt}\Gamma = u X(\Gamma) + v\Theta(\Gamma),\qquad \text{for some } u,v:[0,T]\to\mathbb R
\end{equation}
The new variable $\theta(.)$ plays the role of the direction in $P^1$ of the tangent vector to the curve. Here it is natural to require that $u,v\in L^1([0,T])$, i.e., that  $\Gamma\in W^{1,1}([0,T])$.

\begin{rmrk}
Observe that the lift of a planar curve $\gamma:[0,T]\to\mathbb R^2$ is not unique in general: for example, if $\dot\gamma|_{(\tau_1,\tau_2)}\equiv 0$ for $0\le\tau_1<\tau_2\le T$, it is clear that \eqref{eq-contrSR} is satisfied on $(\tau_1,\tau_2)$ as soon as $u|_{(\tau_1,\tau_2)}\equiv 0$, independently of $v$. Nevertheless, the lift is unique in many relevant cases, e.g., whenever $\dot\gamma(t)= 0$ only on a discrete subset of $[0,T]$.  
\end{rmrk}

\begin{definition}
  A \emph{liftable curve} is a planar curve $\gamma:[0,T]\to \mathbb R^2$ such that $\gamma\in W^{1,1}([0,T],\mathbb R^2)$ and there exists a unique $\th(\cdot)\in W^{1,1}([0,T],P^1)$ such that \r{eq-contrSR} holds for some $u,v\in L^1([0,T])$.
\end{definition}

In particular, for liftable curves $\gamma(\cdot) = (x(\cdot),y(\cdot))$ it holds 
\begin{equation}\label{eq:arctan}
  \theta(t) = \arctan\frac{\dot y(t)}{\dot x(t)},\qquad \text{for a.e.\ }t\in[0,T].
\end{equation}

\subsection{The curve reconstruction problem}

Consider an ``interrupted'' liftable curve, that is, assume that $\gamma:[0,a]\cup[b,T]\to\mathbb R^2$ where $0<a<b<T$.
Let us call $\gamma_{\text{in}}=\gamma(a)$, $(x_{\text{in}},y_{\text{in}}):=(x(a),y(a))$, $\gamma_{\text{fin}}=\gamma(b)$, and $(x_{\text{fin}},y_{\text{fin}}):=(x(b),y(b))$. 
Notice that, according to \eqref{eq:arctan} the limits $\theta_{\text{in}}:=\lim_{\tau\uparrow a}\theta(\tau)$ and  $\theta_{\text{fin}}:=\lim_{\tau\downarrow b}\theta(\tau)$ are well defined.
In the following, we describe a method to reconstruct the missing part, based on the model presented above. That is, we are looking for a curve $\tilde\gamma:[a,b]\to\mathbb R^2$ whose support is a ``reasonable'' completion of the support of $\gamma$.

It has been proposed by Petitot  \cite{pinwheel} 
that the visual cortex reconstructs the curve by minimizing the energy necessary to activate orientation columns which are not activated by the curve itself. This is modeled by the minimization of the following cost functional,
\begin{equation}\label{e-5}
  J=\int_a^b \left(u(\tau)^2+ v(\tau)^2\right)~d\tau\to\min.
\end{equation}
Here, $a,b$ are fixed, and the minimum is taken on the set of liftable curves $\tilde \gamma:[a,b]\to\mathbb R^2$ which are solution of (\ref{eq-contrSR}) for some $u,v\in L^1([a,b])$ and satisfying boundary conditions
\begin{equation}\label{eq-boundary}
\tilde\gamma(a) = \gamma_{\text{in}}, \qquad \tilde\gamma(b)=\gamma_{\text{fin}}, \qquad \tilde\theta(a) = \theta_{\text{in}}, \qquad \tilde\theta(b) = \theta_{\text{fin}}.
\end{equation}
Here $\tilde\theta(\cdot)$ is the lift associated with $\tilde\gamma(\cdot)$ via \eqref{eq:arctan}.
Observe that in the cost \eqref{e-5}, $u(\tau)^2$ (resp.\ $v(\tau)^2$) represents the (infinitesimal) energy necessary to activate horizontal (resp.\ vertical) connections.
As pointed out in \cite{Boscain2012a}, a solution to the problem  \eqref{eq-contrSR}, \eqref{e-5}, \eqref{eq-boundary} always exists. 

\begin{figure}
\centering
  \raisebox{-0.5\height}{\includegraphics[width = .3\textwidth]{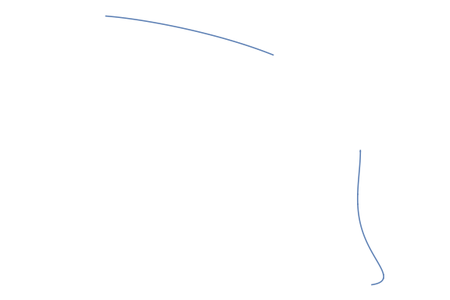}}
  \raisebox{-0.5\height}{\includegraphics[width = .3\textwidth]{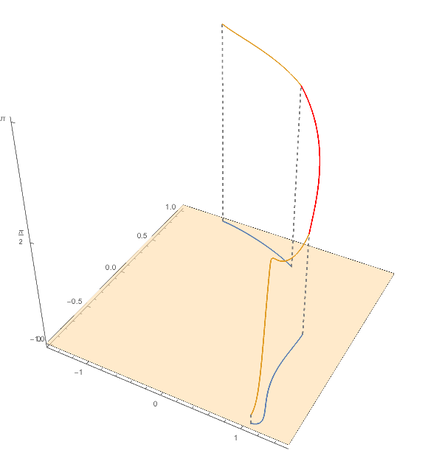}}
  \raisebox{-0.5\height}{\includegraphics[width = .3\textwidth]{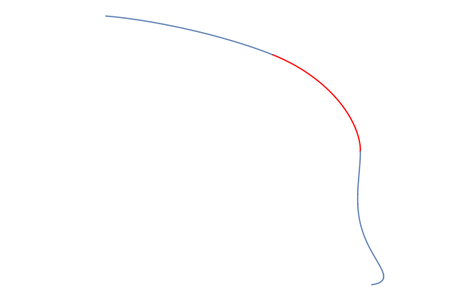}}
  \caption{Curve reconstruction. \emph{From left to right:} The interrupted curve. The lift of the interrupted curve, in yellow, and the solution to the minimization problem, in red. The final reconstructed curve.}
  \label{fig:image-rec}
\end{figure}

An example of curve reconstruction obtained via this technique is given in Figure~\ref{fig:image-rec}.

\brem
 Minimizers of the cost  \eqref{e-5} are minimizers of the reparametrization-invariant cost
\bqn
{\cal L}=\int_a^b \sqrt{ u(\tau)^2+v(\tau)^2}~d\tau=\int_a^b \|\dot{\gamma}(\tau)\| \sqrt{1+\frac{\kappa(\tau)^2}{\beta^2}}~d\tau, \label{e-7}
\eqnl{e-6}
where $\kappa(\cdot)$ is the curvature of the planar curve $\gamma$ and $\beta$ is the dimensional parameter appearing in the vector field $\Theta$. For more details about the relations between minimizers of costs \r{e-5} and \r{e-6}, see \cite{yuri1}. 
\erem

\brem
Observe that here $\theta\in P^1$, i.e., angles are considered modulo $\pi$. Notice that the vector field $X$ is not continuous on $\PTR$. Indeed, a correct definition of problem \eqref{eq-contrSR}, \eqref{e-5}, \eqref{eq-boundary}  needs two charts, as explained in detail in \cite[Remark~12]{Boscain2012a}. In this paper, the use of two charts is implicit, since it plays no crucial role.
\erem
\begin{remark}
\label{remALFA}
From the theoretical point of view, the weight parameter $\beta$ is irrelevant: for any $\beta >0$ there exists a homothety of the $(x,y)$-plane that maps minimizers of problem \eqref{eq-contrSR}, \eqref{e-5}, \eqref{eq-boundary} to those of the same problem with $\beta=1$. However its role will be important in our inpainting algorithms (see Section~\ref{sys-stoch}).
\end{remark}

The minimization problem  \eqref{eq-contrSR}, \eqref{e-5}, \eqref{eq-boundary} is a particular case of a minimization problem called  a sub-Riemannian problem.  For more details on this interpretation see the original papers \cite{citti-sarti,petitot,pinwheel}, the book \cite{petitot-libro}, or 
\cite{Boscain2012a,Boscain2014,nostrolibro} for a language more consistent with the one of this paper.

One the main interests of the sub-Riemannian problem  \eqref{eq-contrSR}, \eqref{e-5}, \eqref{eq-boundary} is the possibility of associating to it a hypoelliptic diffusion equation which can be used to reconstruct images (and not just curves),  and for contour enhancement. This point of view was developed in \cite{Boscain2012a,citti-sarti,Duits2008,Duits2010,Duits2010a}, and is the subject of the next section.

\section{Pure hypoelliptic diffusion for image inpainting}\label{s-hypo}

The sub-Riemannian problem
\eqref{eq-contrSR}, \eqref{e-5}, \eqref{eq-boundary} 
described above  was used to reconstruct  images whose level sets are smooth curves by Ardentov, Mashtakov and Sachkov \cite{ardentov}. The technique they developed consists of reconstructing as the missing parts of the level sets of the image by applying the above curve-reconstruction algorithm.
Obviously, beside the fact that level sets in general need not be smooth nor curves, when applying this method to reconstruct images with large corrupted parts, one is faced with the problem of how to put in correspondence the non-corrupted parts of the same level set.

To avoid this problem, it is natural to model the cortical activation induced by an image $f:\mathbb R^2\to [0,1]$ as a function $\mathcal Lf:PT\mathbb R^2\to \mathbb R$, where $\mathcal Lf(x,y,\theta)$ represents the strength of activation of the neuron $(x,y,\theta)\in PT\mathbb R^2$. 
It is also known  that this cortical activation will evolve with time, due to the presence of the horizontal and vertical cortical connections already introduced in Section~\ref{s-petitot}. Then, a natural way to model this evolution is to assume that the activation of a single neuron propagates as a stochastic process $Z_t$ on $PT\mathbb R^2$, which solves the stochastic differential equation associated with system \eqref{eq-contrSR}. That is, letting $u_t$ and $v_t$ be two one-dimensional independent Wiener processes, we have
\begin{equation}\label{sys-stoch}
  d Z_t = u_t X(Z_t) + v_t \Theta(Z_t).
\end{equation}
Under this assumption, the evolution of a cortical activation $\mathcal Lf$ is obtained via the diffusion naturally associated with \eqref{sys-stoch}, that is, 
\begin{equation}
\begin{cases}
\displaystyle\frac{\partial \psi}{\partial t}= \frac{1}{2}\Delta \psi, \\[.75em]
\psi|_{t=0} = \mathcal L f,
\end{cases}
\qquad\text{where}\qquad 
\Delta = X^2 + \Theta^2 = \biggl(\cos(\theta)\frac{\partial}{\partial x}+
\sin(\theta)\frac{\partial}{\partial y}\biggr)^{2} +
\beta^2\frac{\partial^{2}}{\partial\theta^{2}}. \label{contdif}
\end{equation}
Observe that the above diffusion is highly anisotropic, since the operator $\Delta$ only takes into account two of the three possible directions in $PT\mathbb R^2$. Indeed, $\Delta$ is not elliptic. However, due to the fact that the family of vector fields $\{X,\Theta\}$ satisfy H\"ormander condition, the operator $\Delta$ is hypoelliptic.
The use of equation \eqref{contdif} to model the spontaneous evolution of cortical activations was first proposed by Citti and Sarti in \cite{citti-sarti} (although they posed the problem in $SE(2)$, a double covering of $PT\R^2$).

%

\begin{remark} We have the following relationships between the sub-Riemannian problem control problem \eqref{eq-contrSR}, \eqref{e-5}, \eqref{eq-boundary}, and the diffusion equation \eqref{contdif}: 
\begin{itemize}
\item By the Feynman\,--\,Kac formula, integrating Equation~\eqref{contdif},  one expects to reconstruct the most probable missing level curves (among admissible ones).
\item The solution of \eqref{contdif} is strictly related to the solution of  the minimization problem  \eqref{eq-contrSR}, \eqref{e-5}, \eqref{eq-boundary}.
Indeed, a result by L\'eandre \cite{leandre1,leandre2} shows that, letting $p_t$ be the kernel of \eqref{contdif} and $E(\cdot,\cdot)$ be the value function of problem \eqref{eq-contrSR}, \eqref{eq-boundary} with cost \eqref{e-7}, for all $q,p\in PT\mathbb R^2$ it holds
\begin{equation}
  -4t \log(p_t(q,p)) \longrightarrow E(q,p)^2 \quad \text{as } t\to 0.
\end{equation}
\end{itemize}
\end{remark}

\begin{figure}
   \includegraphics[width=.7\textwidth]{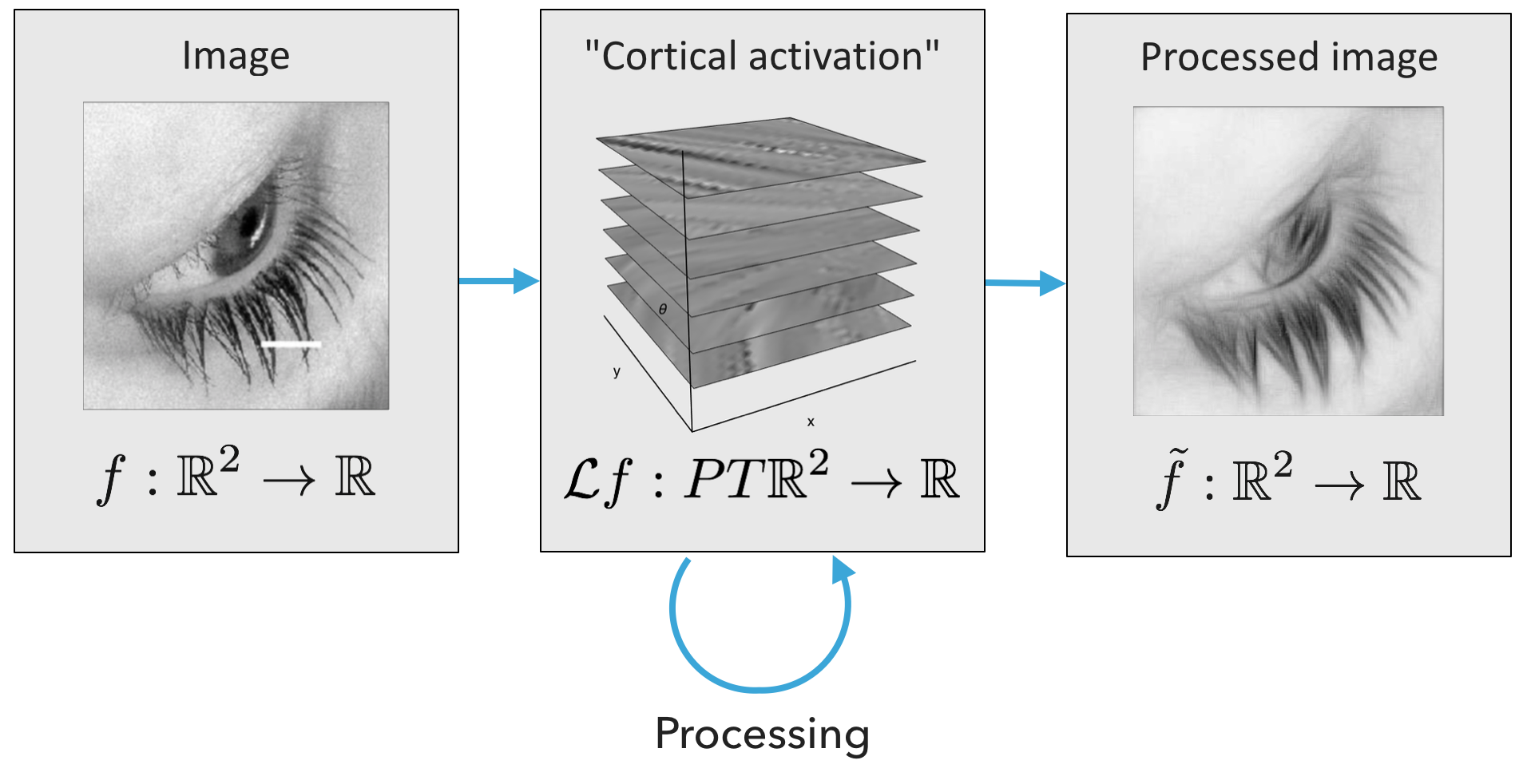}
   \caption{A schematic of the pipeline for image reconstruction.}
   \label{fig:pipeline}
 \end{figure}

To be be able to use equation \eqref{contdif} to reconstruct a corrupted image, one has to specify two things: {\bf i)} the lift operator $\mathcal L$, which maps images on $\mathbb R^2$ to cortical activations on $PT\mathbb R^2$
 {\bf ii)}  how to project the result of the diffusion on $\mathbb{R}^2$ to get the reconstructed image.
(See Figure~\ref{fig:pipeline}.) We remark that in this pipeline we do not exploit any knowledge on the location and/or shape of the corruption.

In the remainder of this section, we will discuss these two points and present an idea of the efficient numerical scheme we exploit to solve \eqref{contdif}.
  
\subsection{Lifting procedure}
Concerning \emph{how} the visual cortex lifts an image to $PT\mathbb R^2$, it seems likely that this is done through multiple convolutions with orientation sensitive filters (like Gabor filters), see e.g. \cite{Daugman1985}. This idea is strictly connected with wavelet transforms and orientation scores, which have been implemented, e.g., in \cite{Duits2008,Duits2010a}. See also \cite{gros-papier}. The main mathematical advantage of this kind of lifts is that, for any $(x,y)\in\mathbb R^2$, the vertical components $\mathcal Lf (x,y,\cdot)$ are essentially 1D gaussians centered around the orientation $\theta(x,y)$ of the gradient $\nabla f(x,y)$.

\begin{figure}
  \includegraphics[width = .4\textwidth]{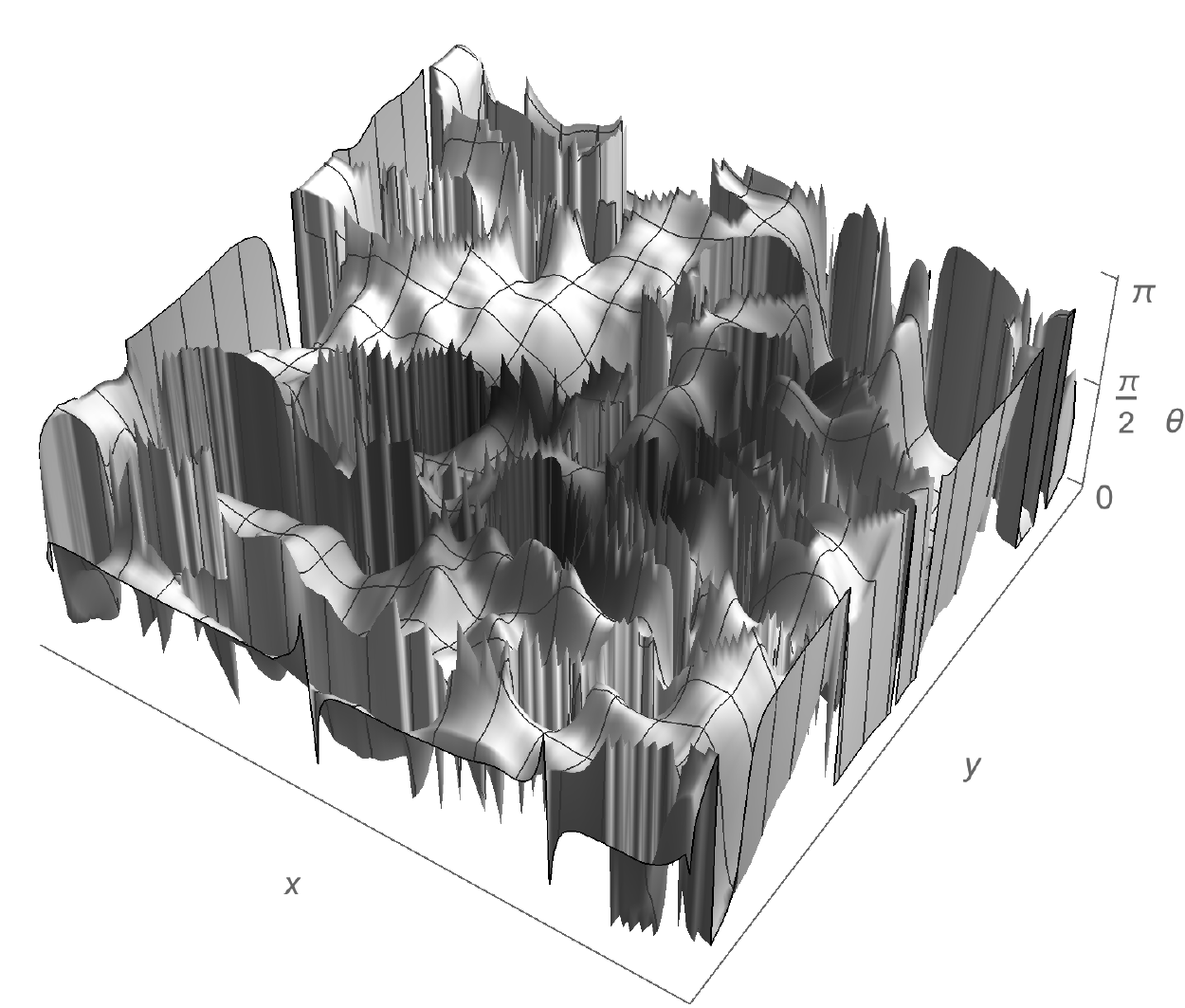}
  \caption{Distributional lift of the image on the l.h.s.\ of Figure~\ref{fig:pipeline}.}
  \label{fig:distr-lift}
\end{figure}

In this paper, we will consider only a primitive version of this lift. (See Figure~\ref{fig:distr-lift}.) Indeed, we will consider the lift $\mathcal Lf:PT\mathbb R^2\to \mathbb R$ of a function $f:\mathbb R^2\to\mathbb R$ to be the distribution $\mathcal L f = f \delta_{Sf}$, where $\delta_{Sf}$ is the Dirac delta measure concentrated on the surface 
\begin{equation}
  Sf = \left\{ \left(x,y,\theta\right)\mid (x,y)\in \mathbb R^2,\, \theta = \theta(x,y) \right\}\subset PT\mathbb R^2 {,\quad 
  \theta(x,y) = \begin{cases}
    \arctan \frac{\partial_y f(x,y)}{\partial_x f(x,y)}, &\quad \text{if } \nabla f(x,y)\neq 0,\\
    P^1 & \quad \text{otherwise.}
  \end{cases}}.
\end{equation}

In order to insure that $Sf$ be a well-defined hypersurface of $PT\mathbb R^2$, we actually compute it on a smoothed version of $f$, see \cite{Boscain2012a}.  This is in accordance with neurophysiological observations that the retina itself operates a Gaussian smoothing.  It is interesting to notice that this smoothing renders $Sf$ particularly regular. More precisely, in \cite{Boscain2012a} it has been shown that the convolution of an $L^2(\mathbb R^2)$ function with a Gaussian is generically a Morse function and that the following holds.

\begin{proposition}[\cite{Boscain2012a}]
  If $f:\mathbb R^2\to \mathbb R$ is a Morse function, then $Sf$ is an embedded $2D$ submanifold of $PT\mathbb R^2$.  
\end{proposition}

We remark that this distributional version of the lift can be interpreted as a limit of the wavelet transform lifts, where the variance of the gaussians obtained by the latter for $\mathcal Lf(x,y,\cdot)$ goes to zero.
Variations on the above lifting procedure have been proposed in \cite{citti-sarti,Sanguinetti,Bredies13}. 

\subsection{Projection}

For our choice of lift, as well as for most of the wavelet transform lifts, solutions to the evolution equation \eqref{contdif} for any time $t>0$ do not belong to the range of $\mathcal L$. Thus, it is not possible to exploit the fact that the lifting procedure is injective in order to define a projection. The most natural way to project functions $\psi:PT\mathbb R^2\to \mathbb R^2$ to $\pi\psi:\mathbb R^2\to \mathbb R$ is then to simply choose
\begin{equation}
  \pi \psi (x,y) = \int_{P^1} \psi(x,y,\theta)\,d\theta.
\end{equation}

\subsection{Numerical integration}

Observe that, since the double covering of $PT\mathbb R^2$ is the group of roto-translations $SE(2)=\mathbb R^2\rtimes\mathbb S^1$, and \eqref{contdif} is covariant w.r.t.\ the canonical projection $SE(2)\to PT\mathbb R^2$, it is more practical to solve \eqref{contdif} on the latter. Indeed, this allows to exploit the group structure of $SE(2)$.
As already remarked, the numerical integration of \eqref{contdif} is subtle, since multiscale sub-Riemannian effects are hidden inside\footnote{For instance, one cannot use the nice method presented in \cite{achdou} for the Heisenberg group. Indeed, in $SE(2)$ one cannot build a refinable grid having a subgroup properties since such grid has at most 6 angles and hence cannot be refined, see e.g.,~\cite{brezzloff}.}
and has been approached in different way. For example, in \cite{citti-sarti,Sanguinetti} the authors use a finite difference discretisation of all derivatives while in \cite{Duits2008,Duits2010,Duits2010a} an almost explicit expression for the heat kernel is exploited. 

In \cite{Boscain2012a,Remizov2013,ahe} we presented a sophisticated and highly parallelizable numerical scheme, based on the non-commutative Fourier transform  on a suitable semidiscretization of the group $SE(2)$, i.e., the semidiscrete group of roto-translations $SE(2,N)$ for $N\in\mathbb N$. This is the semi-direct product $SE(2,N) = \mathbb R^2\rtimes \mathbb Z_N$, where $\mathbb Z_N$ is the cyclic group of order $N$ and the action of $n\in\mathbb Z_N$ on $\mathbb R^2$ is given by the rotation $R_{n}$ of angle $2\pi n/N$. As pointed out in \cite{Remizov2013}, considering a discrete number of orientations seems to be in accordance with experimental evidence. Although still an open problem, this is probably connected with topological restrictions given by the the fact that neurons in V1 encode a 3D space while V1 itself, physically, is essentially 2D. 

A way of interpreting the algorithm presented in \cite{Remizov2013}, see also \cite{visionCDC,gros-papier}, is the following: For a given finite subset $\mathcal K$ of $\mathbb R^2$, let $SE(2,N,\mathcal K)$ be the set of $\mathbb C^N$-valued trigonometric polynomials $Q$ on $SE(2,N)$ that read
\begin{equation}
  \label{eq:appol}
  Q = (Q^n)_{n\in\mathbb Z_N},\quad\text{ where }\quad Q^n(x,y) = \sum_{(\lambda_k,\mu_l)\in \mathcal{K}} c_{k,l}^n e^{i(\lambda_k x+\mu_l y)}, \quad r=0,\ldots,N-1.
\end{equation}
Here, $c^r_{k,l}\in\mathbb C$. Then, if we assume that (standard) Fourier transforms of images are compactly supported on $\mathcal K$, their lifts will belong to $SE(2,N,\mathcal K)$.
Equation \eqref{contdif} can be naturally (semi)discretized on $SE(2,N)$, essentially replacing the operator $\Theta^2$ in $\Delta$ with its discretized version $\Lambda_N\in\mathbb R^N\times \mathbb R^N$, and then restricted to $SE(2,N,\mathcal K)$. This yields the completely uncoupled systems of linear ordinary differential equations
\begin{equation}
  \label{eq:split1}
  \frac{d c_{k,l} }{dt} = - 2\pi^2 \textrm{diag}\left( \lambda_k \cos\theta_r + \mu_l \sin \theta_r \right)^2 c_{k,l} + \Lambda_N c_{k,l}, \qquad c_{k,l}\in\mathbb C^N.\\
\end{equation}
Here, $c_{k,l}(t) = (c_{k,l}^0(t),\ldots,c_{k,l}^{N-1}(t))^T$.
These systems are equipped with initial conditions $c_{k,l}(0) = c_{k,l}$, where the latter are the coefficients in \eqref{eq:appol} corresponding to $\mathcal Lf$.

These discretized equations can then be solved through any numerical scheme.
We chose the Crank-Nicolson method, for its good convergence and stability properties.
Let us remark that the operators appearing on the r.h.s.\ of \eqref{eq:split1} are periodic tridiagonal matrices, i.e.\ tridiagonal matrices with non-zero $(1,N)$ and $(N,1)$ elements.
Thus, the linear system appearing at each step of the Crank-Nicolson method can be solved through the Thomas algorithm for periodic tridiagonal matrices, of computational cost $\mathcal{O}(N)$.

\begin{remark}
  \label{rmk:simmetry}
  By \cite[Theorem~2]{Remizov2013}, solving \eqref{eq:split1} for some couple $(\lambda_k,\mu_l)$ is equivalent to solve it for any rotated couple $R_n(\lambda_k,\mu_l)$, associated with $n\in\mathbb Z_N$.
  Thus, if the set $\mathcal K$ is invariant with respect to rotations $R_{n}$, $n\in \mathbb Z_N$, it is indeed sufficient to solve \eqref{eq:split1} for a slice of $\mathcal K$ whose orbit under the rotations covers the whole $\mathcal K$.
\end{remark}

\subsection{Numerical experiments}

When implementing the pure hypoelliptic diffusion, there are essentially 3 parameters to be tuned: the number of angles $N$ for the semi-discretization of the equation on $SE(2,N)$, the weight parameter $\beta$ appearing in the operator $\Delta$, and the total time of diffusion $T$.
Clearly, for performance reasons, one would like to have $N$ as small as possible. Indeed, numerical experiments suggests that it suffices to choose $N=30$, as increasing $N$ beyond this threshold does not affect the resulting image in any visible way. On the other hand, the parameters $\beta$ and $T$ have to be tuned by hand and the optimal choice seems to be deeply sensitive to the image to be treated (and of course to the desired result).

\begin{figure}
  \includegraphics[height=3.2cm]{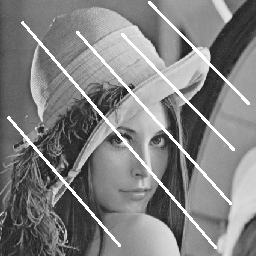}\hfill
  \includegraphics[height=3.2cm]{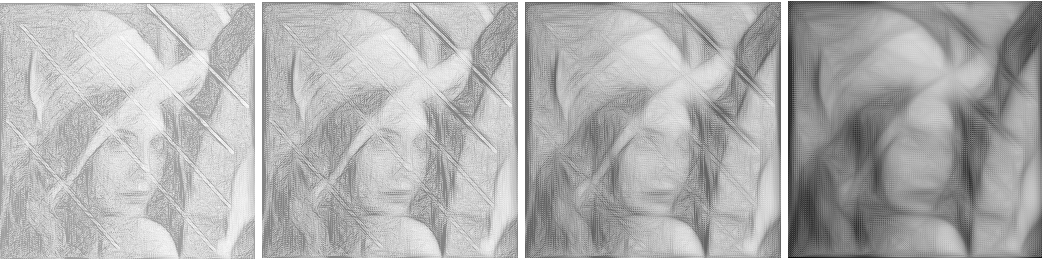}\\ \hfill
  \includegraphics[height=3.16cm]{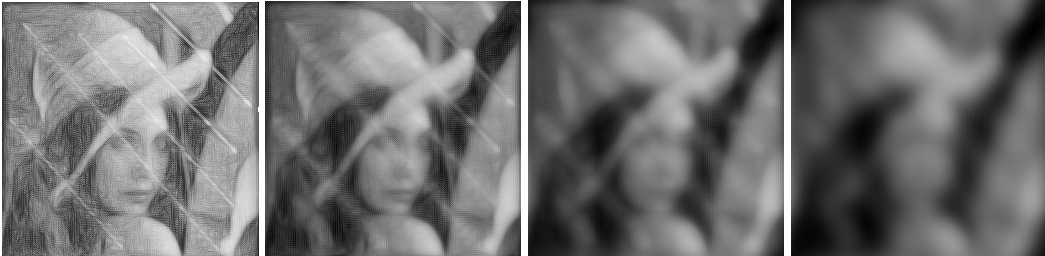}
  \caption{Hypoelliptic diffusion for increasing times $(\frac18,\frac14,\frac12,1)$, from left to right. \emph{Upper left:} original image. \emph{Upper right: }$\beta^2 = \frac14$. \emph{Bottom:} $\beta^2 = 4$.}
  \label{fig:PH}
\end{figure}

\begin{figure}
  \includegraphics[height=3.2cm]{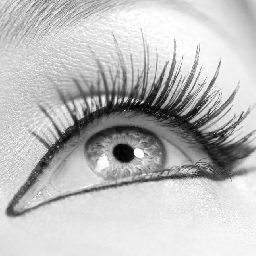}\hfill
  \includegraphics[height=3.2cm]{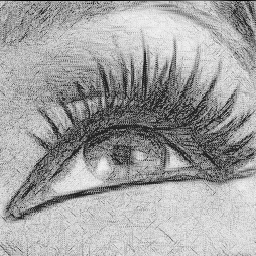}
  \includegraphics[height=3.2cm]{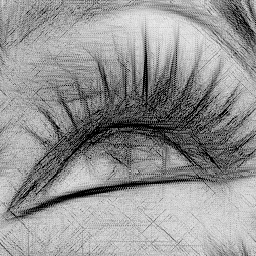}
  \includegraphics[height=3.2cm]{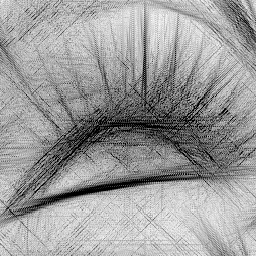}
  \includegraphics[height=3.2cm]{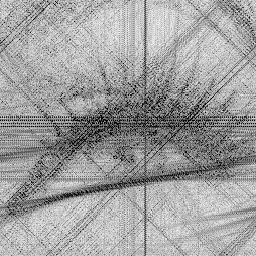}
  \caption{Hypoelliptic diffusion of the image on the left with $\beta = 0$ and increasing times  $(\frac18,\frac14,\frac12,1)$. Notice that, due to the absence of diffusion in the angular part, we observe the formation of straight lines tangent to the level lines.}
  \label{fig:beta0}
\end{figure}

In Figure~\ref{fig:PH}, we present a sequence of images showing the effect of the pure hypoelliptic diffusion with two different choices for $\beta$ at different times.
In particular, this shows how for large values of $\beta$, the effect of the pure hypoelliptic diffusion on $PT\mathbb R^2$ becomes, after the projection, close to an isotropic diffusion on $\mathbb R^2$. 
In Figure~\ref{fig:beta0}, we present the results of the pure hypoelliptic diffusion when the parameter $\beta$ is chosen to be $0$. Notice that, due to the absence of diffusion in the angular part, we observe the formation of straight lines tangent to the level lines.

\begin{figure}
  \includegraphics[height=3.5cm]{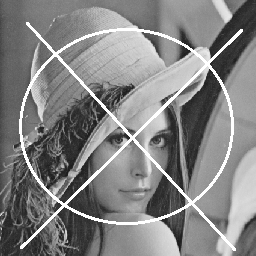}\qquad
  \includegraphics[height=3.5cm]{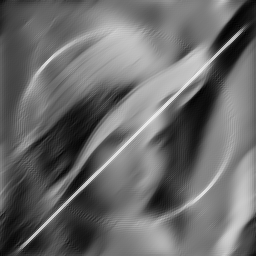}\qquad
  \includegraphics[height=3.5cm]{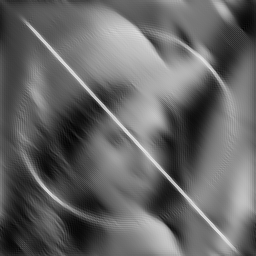}
  \caption{Hypoelliptic diffusion of lift with fixed angle $\theta_0$. \emph{Left:} original image. \emph{Middle:}  $\theta_0=\frac{\pi}4$. \emph{Right:} $\theta_0=\frac{3\pi}4$. In particular observe that only the white stripe perpendicular to $\theta_0$ is filled, while the other is preserved.  }
  \label{fig:PH-theta0}
\end{figure}

\begin{figure}
  \includegraphics[height=3.5cm]{imgs/eye}\qquad
  \includegraphics[height=3.5cm]{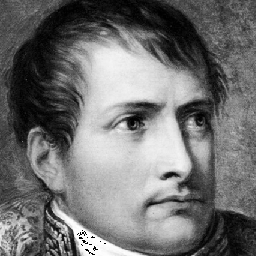}\qquad
  \includegraphics[height=3.5cm]{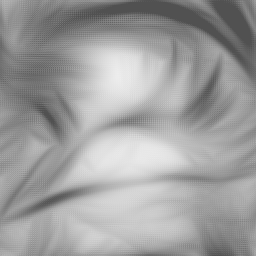}
  \caption{Hypoelliptic diffusion with lift $\tilde{ L}_f g = \delta_{Sf}g$, where $f$ and $g$ are, respectively, the first and second images. Observe that we recover the contours of the first image only through the surface $Sf$. Here we chose $\beta^2 = 0.1$ and $T=1$.}
  \label{fig:PH-no-val}
\end{figure}

In order to show the strong anisotropicity of the hypoelliptic diffusion, we present also Figures~\ref{fig:PH-theta0}, \ref{fig:beta0} \ref{fig:PH-no-val}.
In the first one, we show the effects of the evolution when the surface $Sf$ on which the image is usually lifted is replaced with $S_{\theta_0} = \{ (x,y,\theta_0)\mid (x,y)\in\mathbb R^2\}$, for different choices of $\theta_0\in P^1$. In accordance with the mathematical formulation, this yields a diffusion restricted to the direction $\theta_0$.
Finally, in Figure~\ref{fig:PH-no-val}, we show the hypoelliptic evolution when the lift is replaced with $\tilde L_f g = \delta_{Sf}g$ for two different images $f,g$. Namely, we use the values of the first image only to compute $Sf$ and on this surface we consider the values given by the second image. Here, although the information on the values of the first image is lost, the information on the directions of the level lines alone allows to recover some of the contours.


\section{Heuristic complements: Exploiting information on the location of the corruption}\label{s-corr}

The method explained in the previous section does not use any information of where the image is corrupted. In this section we show how the suitable use of this information permits to obtain better reconstructions. In particular we will present two extensions to the previous algorithm, the Dynamic Restoration (DR) procedure \cite{Remizov2013} and the varying coefficients hypoelliptic diffusion, and briefly introduce a sort of synthesis of the two, the Averaging and Hypoelliptic Evolution (AHE) algorithm \cite{ahe}. The results obtained by these methods are comparable with the current state-of-the-art for PDE based image inpainting algorithms \cite{facciolo,cao}.
  Actually, in our opinion, in order to go beyond this state-of-the-art it is necessary to introduce additional external information on the corrupted part as it is done, for instance, in exemplar-based methods \cite{cao}.

\subsection{Dynamic Restoration}

In this section we present a technique to exploit the information on the location of the corruption in the inpainting algorithm. Assume that a partition of the set of pixels of the image is given $I = G \cup B$ , where points in $G$ are ``good'', i.e., non-corrupted, while those in $B$ are ``bad'', i.e., corrupted. The idea is now to periodically ``mix'' the solution $\psi_t$ of the diffusion on $SE(2,N)$ with the initial function $\mathcal Lf$ on $G$, while keeping tabs on the ``evolution'' of the set of good points. 

Namely, fix $n\in{\mathbb N}$ and split the segment $[0,T]$ into $n$ intervals $t_r=r\tau$, $r=0,\ldots,n$, $\tau=T/n$. Let $G(0)=G$, $B(0)=B$ and iteratively solve  the hypoelliptic diffusion equation on each $[t_r,t_{r+1}]$ with initial condition 
\begin{equation}
  \psi_{t_r}(k,x,y)=
  \begin{cases}
    \psi^-_{t_r}(k,x,y)& \quad\text{if } (x,y)\in B(r)\\
    \sigma(x,y,t_k)\psi^-_{t_r}(k,x,y)& \quad\text{if } (x,y)\in G(r).
  \end{cases}
\end{equation}
Here, the function $\psi^-$ is the solution of the diffusion on the previous interval (or the starting lifted function if $r=0$), and the coefficient $\sigma$ is given by 
\begin{equation}
  \sigma(x,y,t_r)=\frac12 \frac{h(x,y,0)+h(x,y,t_r)}{h(x,y,t_r)}, \qquad h(x,y,t) = \max_k \psi_t(k,x,y).
\end{equation}
Moreover, after each step, $G(r+1)$ and $B(r+1)$ are obtained from $G(r)$ and $B(r)$ as follows:
\begin{enumerate}
  \item[1.] Project the solution $\psi_{t_{r+1}}$ to the image $f_{r+1}$.
  \item[2.] Define $\operatorname{avg}f_{r+1}(x,y)$ as the average of $f_{r+1}$ on the $9$-pixels neighborhood of $(x,y)$.
  \item[3.] Define the set $W = \{(x,y)\in\partial B(r) \mid  f_{r+1}(x,y)\ge \operatorname{avg}f_{r+1}(x,y) \}$.
  \item[4.] Let $G(r+1) = B(r)\cup W$, $B(r+1)=B(r)\setminus W$.
\end{enumerate}

\begin{figure}
  \includegraphics[height=3.5cm]{imgs/lena-diag}\qquad
  \includegraphics[height = 3.5cm]{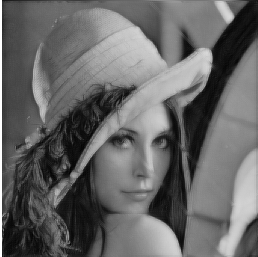}\qquad
  \includegraphics[height = 3.5cm]{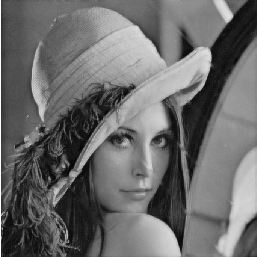}
  \caption{Reconstructions via the DR restoration procedure. \emph{Left: }original image. \emph{Middle:}  $n=30$ steps. \emph{Right: }$n=120$ steps.}
  \label{fig:masking}
\end{figure}

\begin{figure}
  \includegraphics[height = 3.5cm]{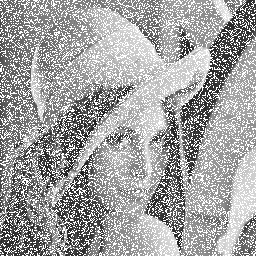}
  \includegraphics[height = 3.5cm]{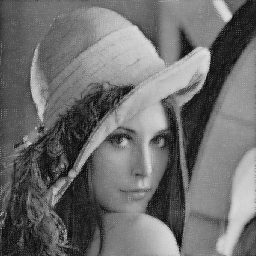}\qquad
  \includegraphics[height = 3.5cm]{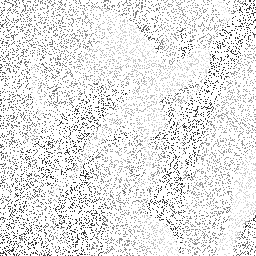}
  \includegraphics[height = 3.5cm]{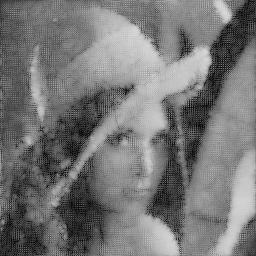}
  \caption{Two inpainting via the DR method of images missing, respectively, 30\% and 80\% of pixels. For both images we chose $T=1$ and $n=60$ steps. Notice how in the second image, the big corruption makes it difficult to correctly compute the gradient.}
  \label{fig:masking-highly}
\end{figure}

Some reconstruction results via the Dynamic Restoration procedure are presented in Figures~\ref{fig:masking} and \ref{fig:masking-highly}. Notice that restorations obtained via this method are much better than those obtained via pure hypoelliptic diffusion. In particular, the DR procedure gives reasonable results even on highly corrupted images, as shown in Figure~\ref{fig:masking-highly}. 

\subsection{Varying coefficients hypoelliptic diffusion}\label{sec:varying}

A different approach w.r.t.\ the Dynamic Restoration procedure is to directly modify the evolution equation \eqref{contdif}, in order to enforce a stronger diffusion on the parts of the image that are known to be corrupted.
Namely, we can replace the operator $\Delta$ in \eqref{contdif} by $\Delta_{\mathcal H} := b X^2 + \frac{a}{\beta^2}\Theta^2$ for some non-negative continuous coefficients $a,b:\mathbb R^2\to \mathbb R$, i.e.,
\begin{equation}\label{eq:varying}
  \Delta_{\mathcal H} =  b(x,y)\biggl(\cos(\theta)\frac{\partial}{\partial x}+
\sin(\theta)\frac{\partial}{\partial y}\biggr)^{2} +
a(x,y)\frac{\partial^{2}}{\partial\theta^{2}}.
\end{equation}
The above equation with varying coefficients $a, b$ tries to implement the natural idea of reducing the eﬀect of diﬀusion at non-corrupted points. Indeed, when using \eqref{contdif}, this can be done only by decreasing of the total time of diffusion $T$, but this acts in an homogeneous way on all points, including the corrupted ones. On the other hand, when using equation \eqref{eq:varying} one can tune the coefficients $a, b$ depending on the point $(x, y)$, and consequently, weaken the diffusion effect only where it is necessary.

Unfortunately, from the point of view of numerical integration, the essential decoupling effect that allowed to reduce \eqref{contdif} to \eqref{eq:split1} does not take place anymore.
In order to overcome this problem, in \cite{ahe}, we approximate the varying coefficient version of \eqref{contdif} by a similar equation with constant coefficients.
Although the constant coefficients operator used in this approximation presents a drift term, this allows to recover the decoupling effect and the Crank-Nicolson method is still pertinent applied to each of the decoupled ODE's.

When choosing the varying coefficients $a$ and $b$, the idea is to make them larger at bad points and their neighbors (especially the coefficient $b$, which has the most influence to the velocity of the diffusion). Thus,  they will be chosen to be a smooth approximation of the indicator function of the the set $B$ of bad (corrupted) points  and $c_0, c_1$ are positive constants. In particular, our choice is:
\begin{equation}\label{9}
  a(x,y) = \begin{cases}
    a_0 + a_1 \epsilon(x,y) & \text{if } \frac{a_0+a_1\epsilon(x,y)}{a_0+a_1}>\epsilon_*,\\
    0& \text{otherwise,}
  \end{cases}\qquad
  b(x,y) = \begin{cases}
    b_0 + b_1 \epsilon(x,y) &  \text{if } \frac{b_0+b_1\epsilon(x,y)}{b_0+b_1}>\epsilon_*,\\
    0& \text{otherwise,}
  \end{cases}
\end{equation}
where
\begin{equation}
  \epsilon(x,y) = \exp \biggl(-\frac{f^2(x,y)}{\sigma}\biggr).
\end{equation}
Here, $a_0, b_0\in\mathbb R$, $a_1,b_1,\sigma>0$, and $\epsilon_*\in(0,1)$ are constant parameters chosen experimentally. 

\begin{figure}
  \includegraphics[height=3.5cm]{imgs/lena-diag}
  \includegraphics[height = 3.5cm]{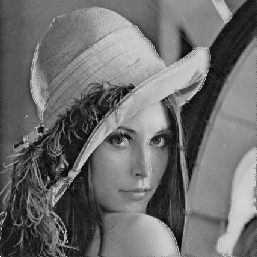}\qquad
  \includegraphics[height=3.5cm]{imgs/lena-random-30}
  \includegraphics[height = 3.5cm]{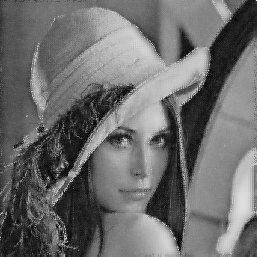}\\ \vspace{.5em}
    \includegraphics[height=3.5cm]{imgs/lena-random-80}
  \includegraphics[height = 3.5cm]{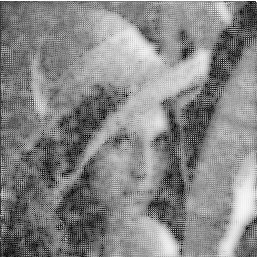}\qquad
  \caption{Three reconstructions via the the varying coefficients restoration procedure: with small diagonal corruptions, and with 30\% and 80\% of pixels randomly removed. Compare with Figures~\ref{fig:PH} and \ref{fig:masking}.}
  \label{fig:varying}
\end{figure}

Some numerical experiments are presented in Figure~\ref{fig:varying}
We observe that for low levels of corruption, reconstruction of images with this method yields results which are comparable to, if not better than, the ones obtained through the DR procedure.
However, when the corrupted part becomes larger this method fails.
This suggests that, in order to obtain a good inpainting algorithm for highly corrupted images, one has still to use the DR procedure, combining it with the varying coefficients. 
This is an essential component of the algorithm presented in the next section.

\subsection{AHE algorithm}

\begin{figure*}[t]
	\begin{center}
\includegraphics[width=\textwidth]{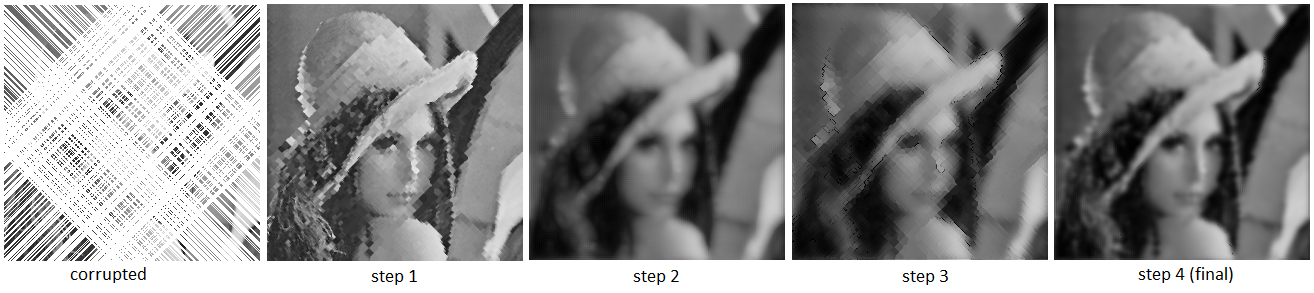}
\caption{A depiction of the different steps of the AHE procedure. Observe how the sharpening from step $2$ to step $4$.}
\label{fig:final-steps}
\end{center}
\end{figure*}

In order to improve on the results for high corruption rates, in \cite{ahe} we proposed the Averaging and Hypoelliptic Evolution algorithm.
The main idea behind the AHE algorithm is to try to provide the anisotropic diffusion with better initial conditions.
More precisely, it is divided in the following 4 steps (see Figure~\ref{fig:final-steps}):
\begin{description}
	\item [1. Preprocessing phase (Simple averaging)] We apply a simple iterative procedure that fills in the corrupted area by assigning to each corrupted pixel the average value of the non-corrupted neighboring pixels.
	\item [2. Main diffusion (Strong smoothing)] By using the result of the previous procedure as an input, we apply the varying coefficients hypoelliptic diffusion discussed in Section~\ref{sec:varying}.
	\item [3. Advanced averaging] In order to remove the blur introduced by the hypoelliptic diffusion of the previous step, we mix the results of step~1 and step~2.
	\item [4. Weak smoothing] We perform a last hypoelliptic evolution, in order to smooth some of the edges we obtained in step~3.
\end{description}

\begin{figure*}[t]
\begin{center}
  \includegraphics[height = 3.5cm]{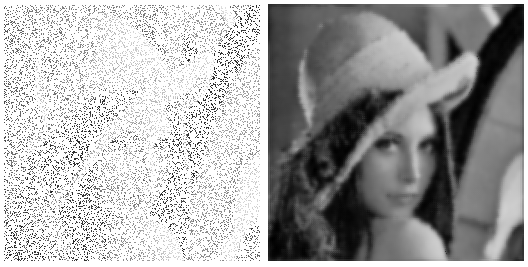}\qquad
  \includegraphics[height = 3.5cm]{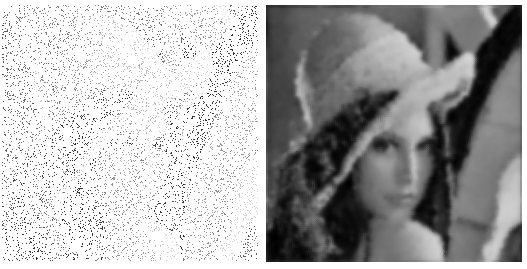}
\caption{Image inpainting via the AHE algorithm of highly corrupted images with, respectively, 80\% and 90\% of pixels missing. Compare with Figures~\ref{fig:masking} and \ref{fig:varying}.}
\label{fig:ahe-highly}
\end{center}
\end{figure*}

In Figure~\ref{fig:ahe-highly} we present the results obtained via the AHE algorithm on highly corrupted images. In particular, a comparison with Figure~\ref{fig:masking-highly} shows that the synthesis between the DR procedure and the varying coefficients, coupled with the averaging steps, yields much better results w.r.t.\ the simple application of the DR procedure.

\section{Conclusions}

\begin{figure}
  \includegraphics[height = 3cm]{imgs/napo-orig}
  \includegraphics[height = 3cm]{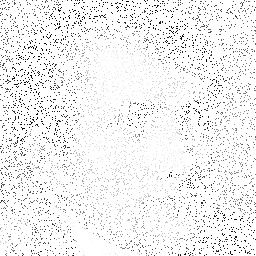}
  \includegraphics[height = 3cm]{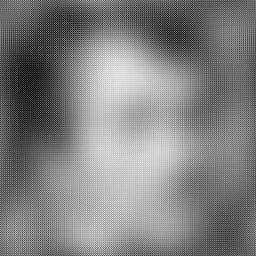}
  \includegraphics[height = 3cm]{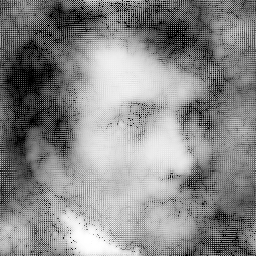}
  \includegraphics[height = 3cm]{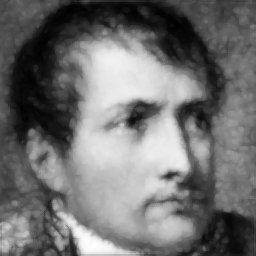}
  \caption{A comparison of the different algorithms presented in this paper. \emph{From left to right:} the original image, the corrupted image used as input, the reconstructions obtained via pure hypoelliptic diffusion ($\beta^2=\frac14,\, T=1$), the DR procedure ($n=120$), and the AHE algorithm.}
  \label{fig:comparison}
\end{figure}

In this paper we presented several image inpainting algorithms based on hypoelliptic diffusion. Although pure hypoelliptic diffusion allows to obtain reasonable inpaints of images with small corrupted regions, we showed that coupling such diffusion with heuristic methods exploiting the full knowledge of the location of the corruption allows to obtain very efficient reconstructions. It is interesting to notice that when the image is so corrupted that our visual system is not able to recognize it, as it happens, for instance, for the corrupted image presented in Figure~\ref{fig:comparison}, the use of pure hypoelliptic diffusion does not help it. This fact can be interpreted as a validation of the CPS model with pure hypoelliptic diffusion. On the contrary, as already pointed out, both the DR procedure and the AHE algorithm produce reconstructions that go beyond the capabilities of our visual system.  

\section*{Acknowledgments}
This work was supported  the ERC POC project ARTIV1, project number
727283, by the ANR project SRGI “Sub-Riemannian Geometry and Interactions”,
contract number ANR-15-CE40-0018, and by a public grant as part of the ``Investissement d'avenir project'', reference ANR-11-LABX-0056-LMH, LabEx LMH, in a joint call with the “FMJH Program Gaspard Monge in optimization and operation research”.
The second author was supported by the project POCI-01-0145-FEDER-006933/SYSTEC financed by ERDF through COMPETE2020 and by FCT(Portugal). The third and fourth authors have been supported by the CNRS INS2I PEPS project CINCIN.

\bibliographystyle{abbrv}

\end{document}